%% file: main.tex
\documentclass[10pt,twocolumn,letterpaper]{article}

\usepackage{cvpr}              

\usepackage{multirow}
\usepackage{adjustbox}
\usepackage[dvipsnames]{xcolor}
\usepackage{pifont}
\newcommand{\cmark}{\textcolor{Green}{\ding{51}}}%
\newcommand{\xmark}{\textcolor{Red}{\ding{55}}}%
\usepackage{svg}
\usepackage{colortbl}
\usepackage{xcolor}
\usepackage{comment}

\definecolor{cvprblue}{rgb}{0.21,0.49,0.74}
\usepackage[pagebackref,breaklinks,colorlinks,allcolors=cvprblue]{hyperref}

\def\paperID{18736} 
\def\confName{CVPR}
\def\confYear{2025}

\title{Cross-Modal Distillation for 2D/3D Multi-Object Discovery from 2D motion}

\author{
  \centering
  Saad Lahlali \footnote{Equal contribution.}\and Sandra Kara\footnotemark[1] \and Hejer Ammar  \and Florian Chabot \and Nicolas Granger \and Hervé Le Borgne \and Quoc-Cuong Pham \and
  Université Paris-Saclay, CEA, List, F-91120, Palaiseau, France\\
  {\tt\small {firstname.lastname}@cea.fr}
}

\usepackage[normalem]{ulem}
\usepackage{tabularray}

\begin{document}
\maketitle
\input{sec/0_abstract}    
\input{sec/1_intro}
\input{sec/2_related_works}
\input{sec/3_method}
\input{sec/4_experiments}
\input{sec/5_conclusion}


{
    \small
    \bibliographystyle{ieeenat_fullname}
    \bibliography{main}
}
\input{supp_mat}
{
    \small
    \bibliographystyle{ieeenat_fullname}
    \bibliography{main}
}
\end{document}

%% file: sec/0_abstract.tex
\begin{abstract}
Object discovery, which refers to the task of localizing objects without human annotations, has gained significant attention in 2D image analysis. However, despite this growing interest, it remains under-explored in 3D data, where approaches rely exclusively on 3D motion, despite its several challenges. In this paper, we present a novel framework that leverages advances in 2D object discovery which are based on 2D motion to exploit the advantages of such motion cues being more flexible and generalizable and to bridge the gap between 2D and 3D modalities.
Our primary contributions are twofold: (i) we introduce DIOD-3D, the first baseline for multi-object discovery in 3D data using 2D motion, incorporating scene completion as an auxiliary task to enable dense object localization from sparse input data; (ii) we develop xMOD, a cross-modal training framework that integrates 2D and 3D data while always using 2D motion cues. xMOD employs a teacher-student training paradigm across the two modalities to mitigate confirmation bias by leveraging the domain gap. During inference, the model supports both RGB-only and point cloud-only inputs. Additionally, we propose a late-fusion technique tailored to our pipeline that further enhances performance when both modalities are available at inference.
We evaluate our approach extensively on synthetic (TRIP-PD) and challenging real-world datasets (KITTI and Waymo). Notably, our approach yields a substantial performance improvement compared with the 2D object discovery state-of-the-art on all datasets with gains ranging from $+8.7$ to $+15.1$ in $F1@50$ score.  The code is available at \url{https://github.com/CEA-LIST/xMOD}


\end{abstract}

%% file: sec/1_intro.tex
\section{Introduction}
\label{sec:intro}
Object detection has been extensively explored, leading to fast, high-performance approaches \cite{faster,yolo,detr}. However, these methods adopt a fully supervised setting that suffers from high annotation costs and makes them impractical for scaling with the increasing data needed for better generalization. Additionally, this setting is limited to detecting specific semantic categories, which poses challenges in identifying out-of-distribution instances and rare categories. Object discovery has thus emerged as an unsupervised alternative to the \textit{localization} component of object detection. It focuses on localizing \textit{objects} within images or videos without explicit prior knowledge provided by human annotations. Interest in this task continues to grow in the 2D modality \cite{motok,dinosaur,safadoust,DIOD} driven by the presence of object patterns \textit{for free} within low-level and automatically acquired modalities (motion \cite{DOM, SAVI}, depth \cite{SAVIpp}, etc), resulting in interesting performances. Moreover, the class-agnostic nature of object discovery and its reliance on low-level signals allow for a broader application, built around general definitions of objects, such as salient objects \cite{LOST,TokenCut} and objects that can move \cite{DOM}. These properties address the limitations of the fully supervised setting. In contrast, these advances are not mirrored enough in the 3D modality where only 3D motion cues are explored despite being sparse and demanding extensive fine-tuning with changing domains. 


In this work, we show that 3D object discovery (3DOD) can largely benefit from advancements achieved in the 2D modality. Specifically, we adapt the recent motion-guided 2D object discovery (2DOD) approach in \cite{DIOD} to accommodate 3D data, using the same 2D motion masks. Object discovery being unsupervised, it typically includes a reconstruction pretext task as a powerful regularization method. In the real-world scenario, we discovered that the inherent sparsity in LiDAR data (\ie missing data points and poor spatial resolution) makes 3DOD challenging, leading to incomplete object segments. We thus propose scene completion as a more suitable pretext task for 3DOD. Specifically, we encourage the prediction of a denser point cloud, which helps avoid propagating the input sparsity to the predicted object masks.

Subsequently, we aim to ensure that the transition to 3D is not disconnected from the 2D data, which is rich in complementary information such as colors and textures. To this end, we propose bridging the two modalities by jointly optimizing the tasks of 2D and 3D object discovery, while always using the same 2D motion cues. The effectiveness of distillation for object discovery has been demonstrated in an intra-modal setting \cite{DIOD}, where it progressively reintegrates discovered objects into the supervision set and eliminates noisy pseudo-labels, enhancing robustness. In our work, we explore distillation in a cross-modal setting. To achieve this, we design two teacher-student systems, one for each modality, and establish interactions between the four models using objective functions that enable the student model of one domain to be supervised by the teacher model of the alternate domain. Advantageously, our approach increases the robustness of the system when a modality becomes inoperative due to a difficult environment, such as night scenes for a camera (\textit{2D-blind}) or the absence of reflections for a 3D sensor (\textit{3D-blind}). This process also leverages the domain gap between the student and teacher models, as each receives inputs from a distinct modality, reducing the risk of confirmation bias.

During inference, our method can accommodate 2D only, 3D only and multi-modal inputs, depending on the application and available sensors. In the multi-modal setting, we explore the consistency between both modalities as a source of reliability in multi-sensor applications, considering consistent predictions between the two modalities as the most reliable object candidates.

In summary, (i) we propose a first baseline to solve multiple object discovery from point clouds using 2D motion cues, with scene completion as a suitable pretext task for 3DOD; (ii) we design a cross-modal training framework, based on 2D motion information, that integrates 2D and 3D data to enable interaction between the two modalities, addressing modality-related difficult cases. Experiments conducted on three datasets demonstrate that each modality benefits significantly from cross-modal learning with the alternate modality, validating the effectiveness of the proposed approach. 


%% file: sec/2_related_works.tex
\section{Related Work}
\label{sec:Related Work}

\subsection{Object discovery in RGB images (2DOD)}
\label{sec:2DOD}
Object discovery in 2D images/videos addresses the challenge of localizing instances of objects when human annotations are unavailable. In RGB images, this task has significantly benefited from advances in self-supervised learning \cite{DINO, dinov2}, which have led to the emergence of segmentation properties in learned representations \cite{survey}. Notably, DINOSAUR \cite{dinosaur} demonstrated that reconstructing those pre-learned features enables self-supervised scene decomposition into objects.

Recently, 2DOD has achieved greater success in video data, driven by the availability of motion information that serves as a cue for object localization. Motion information has been primarily incorporated into slot-attention-based approaches, with slot-attention being the mechanism that  facilitates scene decomposition into objects within the latent space of an auto-encoder architecture \cite{SA}. Motion is integrated in various ways across different methods: SAVI \cite{SAVI} learns to predict optical flow, focusing particularly on the localization of moving objects. On the other hand, VideoSAUR \cite{videosaur}, a video version of \cite{dinosaur}, exploits semantic similarity between image patches to predict their temporal displacement, thus incorporating motion implicitly into the learned representation. More explicitly, another research direction \cite{DOM,motok,BMOD,DIOD} leverages motion-derived segments, highlighting moving objects, to guide slots’ learning; some approaches also address noise in image backgrounds \cite{BMOD} and the generalization from moving to static objects \cite{DIOD}. 



Although these methods demonstrated interesting results for 2DOD, these advances along with the use of 2D motion cues have not been exploited yet for both 3DOD and corss-modal object discovery.

\subsection{Object Discovery in 3D data (3DOD)}


Compared to 2DOD, 3DOD is less explored \cite{paper7,paper5}. In single images, it is typically limited to single-object localization \cite{wang2023autorecon}, while in sequential data, the primary approach leverages 3D motion cues to identify only moving objects \cite{paper2,paper5}. However, in LiDAR-based applications like road scenes, ignoring stationary objects, such as stopped vehicles, raises safety concerns.
In an other category, Open-set detection \cite{paper3} generalizes to unknown objects but primarily relies on highly-supervised closed-set detectors, while vision-language methods \cite{paper11} assume known or  describable classes of objects, which is more restrictive than general object discovery.
Other approaches \cite{wang20224d, paper6, paper7, paper8}, while unsupervised, mostly cluster 3D point clouds \cite{paper6,paper7} or scene flow cues \cite{wang20224d}, requiring intensive tuning and heuristic-based filtering of non-object regions \cite{paper6}. Clustering in 3D data is further challenged by LiDAR's low resolution and sparse points on distant objects.

In this work, we aim to extend advancements from 2DOD (Section \ref{sec:2DOD}) to the 3D domain. Similar to how 2D object-centric learning offers a deep learning-based alternative to 2D clustering, this extension seeks to replace clustering methods for 3D point clouds, which are sensitive to parameters like object count and intra-object point density. Our hypothesis is also that 3D data can, in turn, enhance object discovery in 2D, thus the proposed cross-modal distillation framework.

\subsection{Motion Cues for Object Localization}
\label{OF_vs_SF}
 An important part of understanding a scene is modelling its dynamics. This has motivated many works on motion estimation both in RGB images through optical flow estimation (\ie. the pixel displacement between successive frames) \cite{PWC,RAFT,SEARAFT} and in 3D by estimating 3D displacements of each point, known as scene flow \cite{FN3D,MIT,ICP}.
Motion information has notably served as a cue for the presence of objects of interest: moving objects \cite{TSAM}, objects capable of moving \cite{DOM}, \etc
For instance, in \cite{semoli} which addresses semi-supervised segmentation of moving objects in point clouds, scene flow is employed to localize mobile objects. Conversely, 2D methods utilize optical flow for scene analysis \cite{motionGroup,layerSeg}.

 
In this work, the choice of using 2D-derived motion cues, instead of 3D scene flow offers several advantages: (i) It avoids the need for pre-processing steps like ground removal in point clouds, a common requirement in clustering-based methods \cite{groundRemoval} that entails additional hyper-parameter tuning. Recent advances in video object discovery (\ref{sec:2DOD}) handle this automatically, even filtering out other permanently static regions such as buildings. (ii) Using the 2D domain as the source for pseudo-labels, rather than point clouds, helps reduce errors associated with the low resolution of LiDAR data, particularly on distant objects. (iii) Finally, leveraging 2D-derived supervision to solve 3DOD opens the perspective of using the vast resource of foundation models emerging rapidly in the 2D domain \cite{yoloworld,groundingdino}, and transferring this knowledge into the 3D space.

%% file: sec/3_method.tex
\section{Method}
\label{sec:method}

\begin{figure*}[tb]
  \centering
\includegraphics[trim={1cm 0 1cm 0},clip,width=0.9\linewidth]{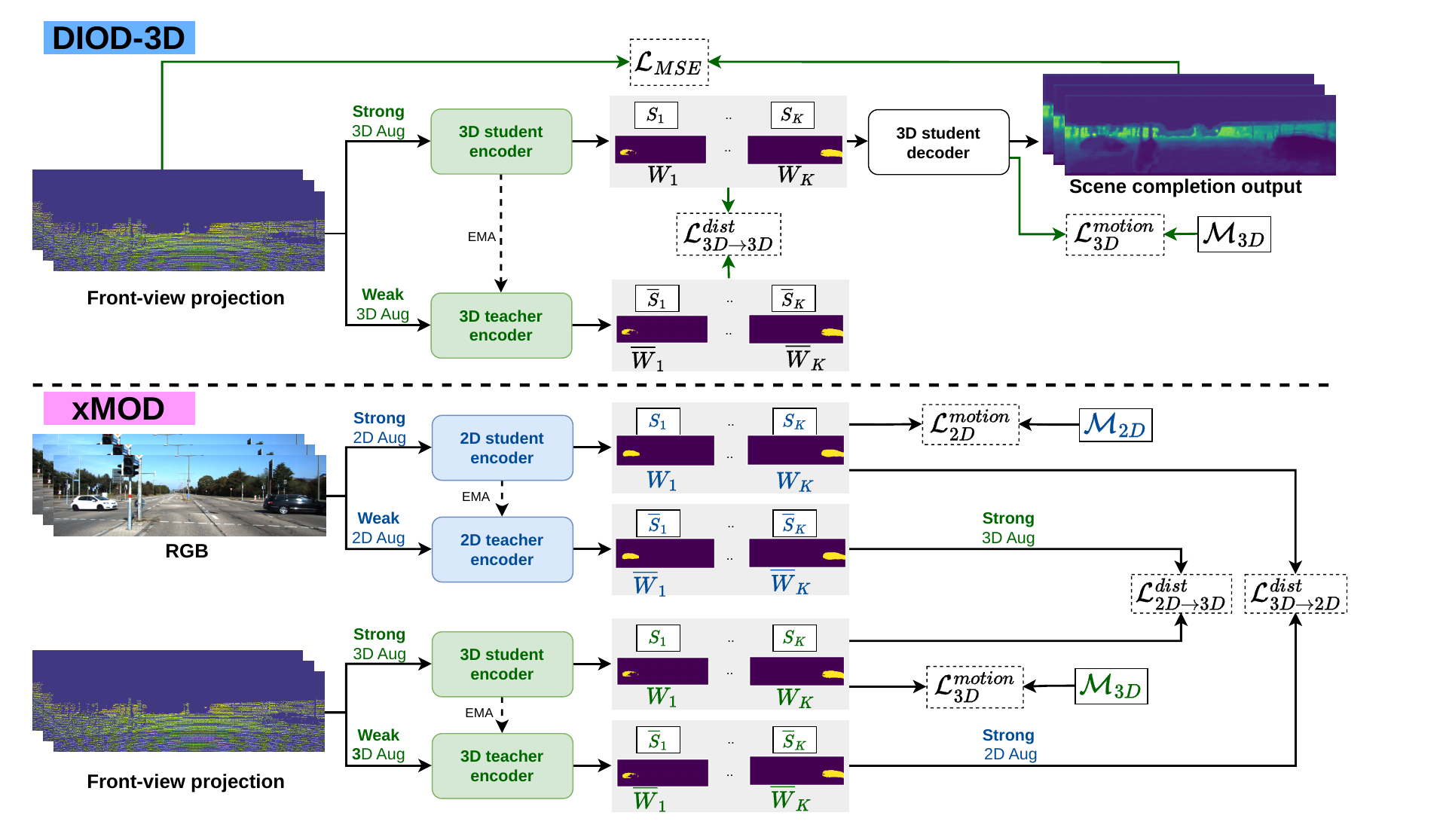}
  
   \caption{\textbf{Overview of the proposed approach}. \textbf{i) DIOD-3D}. At each iteration, a sequence front-view projections of point clouds is passed to the 3D teacher and student models. Attention maps from the teacher model are presented as targets to the student model through $L^{dist}_{3D\rightarrow3D}$.  An MSE objective is employed to predict the original scene from input with missing data, enabling 3D scene completion as an auxiliary task for 3DOD. \textbf{ii) Cross-modal distillation (xMOD)}. Alongside the 3D branch, sequences of RGB images are forwarded to the 2D teacher and student models. $L^{dist}_{2D\rightarrow3D}$ means pseudo-labels from the 2D teacher model are aligned with the 3D student input and used for its supervision; $L^{dist}_{3D\rightarrow2D}$ works similarly for 3D to 2D pseudo-labeling. Motion pseudo-labels $\mathcal{M}_{2D}$ and $\mathcal{M}_{3D}$ are used for regularization, with $\mathcal{M}_{3D}$ being the 2D motion segments with corresponding 3D points. We omit representing 2D reconstruction and 3D completion task for simplification.}
   \label{fig:overview}
\end{figure*}

Our method consists of two main components (\autoref{fig:overview}). First, we introduce an approach for multi-object discovery from 3D data based on 2D motion, which we call DIOD-3D. Next, we design a cross-modal distillation framework (xMOD) that enables interaction between two branches, xMOD (2D) and xMOD (3D), which process 2D and 3D data, respectively, and generate pseudo-labels for the alternate modality.
In the following sections, we first outline the 2DOD method that forms the foundation of our approach, before describing the two distinct aspects of our work.

\subsection{Context: distilled motion-guided slot attention for 2D object discovery}
\label{sec:context}

The objective is to utilize automatically acquired motion information to localize mobile objects; and to generalize to other static objects within the same semantic category \cite{DOM}. A recent approach specifically addresses the challenge of generalization by proposing a method that first uses as targets pseudo-labels for mobile objects, generated from optical flow. Leveraging these pseudo-labels, a distillation framework is employed to gradually expand the pseudo-labels set to include static objects identified by the model, thus covering all instances within the semantic category of interest \cite{DIOD}.
Specifically, during an initial burn-in phase, the model processes a sequence of $T$ frames to generate a video representation  $H^{t} \in  h \times w \times D$ at each timestep $t$. The features $H$ are distributed across $K$ slots (queries) through an attention module in two main steps: i) Attention weights $W$ are computed between $H^t$ and the set of slots from the previous timestep as $W^t = \frac{1}{\sqrt{D}} k(H^{t}) \cdot q(S^{t-1}) \in \mathbb{R}^{N \times K}$, ii) Slots $S^t$ are updated as $ {W^t}^\top v(H^t)$, where $v$, $k$, and $q$ are three learnable projections and $N = h \times w$ \cite{attention}.
To enable objects activation within the attention maps, these are supervised using a set $\mathcal{M}_{2D} = \{ \mathbf{m}_l \in \{0,1\}^{h \times w} : l \in \{1, \dots, L\} \}$ of pseudo-labels extracted from optical flow \cite{DOM}, with $L$ being the number of pseudo labels available for a given image. These masks are aligned with the model-generated attention maps through Hungarian matching. The background class is isolated within a specific attention map $W_{bg}$ using a negative log-likelihood loss function, as described in \cite{BMOD}.
Following the burn-in, the model enters a distillation phase where it is duplicated into teacher and student models. The student model learns to discover objects through gradient back-propagation, while the teacher model is updated as an exponential moving average (EMA) of the student, ensuring gradual refinement of model capabilities. Notably, during the burn-in phase, the teacher model learns to generalize from moving objects to static objects within the same category through semantics. Distillation then allows both moving and static objects extracted from the teacher model to be presented as targets to the student model. Specifically, any connected region in one teacher’s attention map $\overline{W}$ is identified as a candidate object and, if it passes a confidence test, is added to the targets for supervising the student model. For supervision, a weighted Binary Cross-Entropy (BCE) loss function is employed, where weighting is based on the confidence associated with each object segment. Alongside the teacher’s predictions, the motion pseudo-labels continue to be used during the distillation phase and act as a regularization.

\subsection{3D Object Discovery}
\label{sec:method1}

The inherent sparsity in 3D data is a challenge for tasks like object detection, in particular in the unsupervised setting of object discovery, where detailed and complete input information is crucial. Additionally, directly processing raw 3D data requires more complex and computationally intensive algorithms. To address this, 2D projections of point clouds are used to transfer data into a denser grid-structured space, manageable by efficient 2D models. 

\subsubsection{DIOD-3D: our approach for 3D Object Discovery}
\label{sec:3DOD}

For each scene, the corresponding LiDAR-generated point cloud (\ie a set of 3D points) is projected into 2D using a front-view projection, as shown in \autoref{fig:overview}. 
Let $\mathbf{I}_{\text{fv}} \in \mathbb{R}^{H' \times W' \times 4}$ be the projected 2D image matrix for a given scene. Each pixel in $\mathbf{I}_{\text{fv}}$ contains four channels: $ \mathbf{I}_{\text{fv}}(i, j) = (X_{ij}, Y_{ij}, Z_{ij}, d_{ij}) \quad 
\text{for } i \in \{1, \dots, H'\} \text{ and } j \in \{1, \dots, W'\}$, with $d$ being the distance of the projected points from the RGB camera origin. 
The pixel $(i, j)$ is assigned a fill-value vector $(f, f, f, f)$, where $f$ is set to $0$ in this work to indicate the absence of an associated 3D point. This can occur either due to the LiDAR's lower resolution compared to the camera or because the camera's vertical field of view (FOV) is wider than that of the LiDAR.

Due to the inherent differences in the vertical FOV between the LiDAR and RGB camera, the motion pseudo-labels extracted from the optical flow (in the 2D domain) can occupy regions without any corresponding 3D information, particularly at the top of the projected image. This has been observed to cause model hallucinations in those regions, in the form of high-confidence noise segments.
To address this, motion masks without corresponding 3D data are discarded in the motion guidance. We denote the new set of motion pseudo-labels as $\mathcal{M}_{3D}$. For each scene $\mathbf{I}_{\text{fv}}$, $\mathcal{M}_{3D}$ is a subset of the 2D pseudo-labels $\mathcal{M}_{2D}$ defined as: 
\begin{equation}
\small
\mathcal{M}_{\text{3D}} = \left\{
\mathbf{m}_l \in \mathcal{M}_{\text{2D}} \ \middle|\ 
\begin{aligned}
&\exists (i, j) \text{ such that } \mathbf{m}_l(i, j) = 1 \\
&\text{and } (X_{ij}, Y_{ij}, Z_{ij}, d_{ij}) \neq (f, f, f, f)
\end{aligned}
\right\}.
\end{equation}


Let $m_{3D} \in \mathcal{M}_{\text{3D}}$ be a motion pseudo-label for the scene $\mathbf{I}_{\text{fv}}$, that matches the attention map $W$ (Hungarian matching) learned by the student model. Motion supervision is applied via the following BCE loss:

\begin{equation}
\small
    \begin{aligned}
    \mathcal{L}^{motion}_{3D}(m_{3D}, W)&=-\frac{1}{N}\sum_{i=1}^{N} \big[\big(1+s_{m_{3D}}\big)m_{3D}(i)\log\big(W(i)\big) \\
      &+ \big(1-m_{3D}(i)\big)\log\big(1-W(i)\big)\big]
    \end{aligned}
\end{equation}

\noindent where the confidence score $s_{m_{3D}}$ is computed as the average activation within the learned foreground map $W_{fg}$ at the object's location in $m_{3D}$.


Similar to the 2D approach in \cite{DIOD}, $\mathcal{L}^{motion}_{3D}$ is employed as the sole supervisory signal during the burn-in phase. During the distillation phase, each highly confident teacher-generated pseudo-label $c$ is incorporated as a target using $L^{dist}_{3D \rightarrow 3D}(c, W)$ (same definition as $\mathcal{L}^{motion}_{3D}$).

\subsubsection{Scene Completion as a Pretext Task for 3DOD}
\label{sec:SC}

Scene reconstruction has proven to be an effective pretext task in RGB images~\cite{motok}. However, this conclusion does not hold for the task of 3DOD. Trying to reproduce the high and variable sparsity of LiDAR data makes scene understanding challenging, and results in sparse and less accurate predictions. Refer to the ablation study in \autoref{abl} for a quantitative evaluation of these limitations.

For this reason, we propose to rely on scene completion as a pretext task for 3DOD. Let $\mathcal{P}$ be the set of coordinates corresponding to valid projections of 3D points. We randomly remove a subset $ \mathcal{P}_{\text{drop}} \subset \mathcal{P} $ from these coordinates. The objective is then to reconstruct the pixels at positions in $\mathcal{P}$ using the information from pixels at positions in $ \mathcal{P} \setminus \mathcal{P}_{\text{drop}} $ (see \autoref{fig:overview}). The reconstruction is guided by a mean squared error loss, optimized only for valid projections of 3D points to avoid reproducing the input sparsity; and is defined as:
\begin{equation}
\mathcal{L}_{\text{MSE}} = \frac{1}{|\mathcal{P}|} \sum_{(i,j) \in \mathcal{P}} \left( \hat{\mathbf{I}}(i,j) - \mathbf{I}_{\text{fv}}(i,j) \right)^2
\end{equation}
\noindent where $\hat{\mathbf{I}}$ and $\mathbf{I}_{\text{fv}}$ are the reconstructed and original frontal projections. The previous objective enables scene completion behavior, which enhances scene understanding and segmentation.

\subsection{Cross-Modal Distillation for Unsupervised 2D/3D Object Discovery }
\label{sec:cross}

In the previous sections, we proposed a first method for real-world object discovery using LiDAR data. Our approach is based on intra-modal distillation, where both the student and teacher models receive the same 3D data.
Even when these data are augmented differently, the gap between the two inputs remains limited, suggesting that the teacher's contribution to the student might be reduced in this setting. This assumption is confirmed in the ablation study presented in \autoref{abl}. 

In this section, we propose a cross-modal distillation framework that places the teacher and student models in two different domains: 2D and 3D modalities. Specifically, we jointly optimize two teacher-student systems, one in each modality, and enable pseudo-labeling from the teacher in one modality to the student in the alternate modality, as shown in \autoref{fig:overview}. Thus, the 3D teacher provides a guidance signal from the 3D domain, addressing the limitations of the 2D student in \emph{2D-blind} scenarios (such as night scenes or fog). The 2D teacher, in turn, enhances the robustness of the 3D student in \emph{3D-blind} scenarios such as objects with low reflectivity or highly cluttered environments.

Concretely, at each iteration, a video sequence of length $T$ is passed through the four models (2D teacher, 2D student, 3D teacher, and 3D student), with strong modality-specific augmentations applied to the inputs of the student models. Attention maps are produced by the slot-attention module within each model and are involved in the cross-model supervision. The attention maps from the teacher models are binarized to generate object candidates as described in \cite{DIOD}. For simplicity, we will consider the case where $T=1$ frame. Let $D_{1}$ and $D_{2}$ be the source and target domains, respectively, during the exchange of pseudo-labels. We denote $c_{D_{1}}$ an object candidate proposed  by the teacher model of domain $D_{1}$, which matches the attention map $W_{D_{2}}$ of the student model from domain $D_{2}$.
The inter-modal distillation objective function for the previous pair is defined as follows:

\begin{equation}
\small
    \begin{aligned}
    \mathcal{L}^{dist}_{D_{1} \rightarrow D_{2}}(c_{D_{1}}, W_{D_{2}})&=-\frac{1}{N}\sum_{i=1}^{N}\big[\big(1+s_{c}\big)c_{D_{1}}(i)\log\big(W_{D_{2}}(i)\big)\\
      &+ \big(1-c_{D_{1}}(i)\big)\log\big(1-W_{D_{2}}(i)\big)\big]
    \end{aligned}
\end{equation}

\noindent $D_{1}$ and $D_{2}$ can be either 2D or 3D modalities, based on the direction of the pseudo-label exchange.  Specifically:

\begin{itemize}
    \item $\mathcal{L}^{dist}_{2D \rightarrow 3D}(c, W)$ when the object candidate $c$ is derived from the 2D teacher and $W$ is a learned 3D student's attention map.
    \item $\mathcal{L}^{dist}_{3D \rightarrow 2D}(c, W)$ when the object candidate $c$ is derived from the 3D teacher and $W$ is a learned 2D student's attention map.
\end{itemize}

The case where $ D1 = D2 \in \{2D, 3D\} $ corresponds to intra-modal distillation, which is not utilized as an objective in our proposed training approach. The ablation study in section \ref{abl} demonstrates the ineffectiveness of this distillation compared to inter-modal pseudo-labelling.

Given the findings presented in section \ref{sec:SC}, we employ scene completion as a pretext task for the 3D branch, while the 2D branch continues to pursue a 2D scene reconstruction objective. Additionally, the motion masks $\mathcal{M}_{2D}$ and $\mathcal{M}_{3D}$ are still used as targets for the 2D and 3D branches, respectively, for regularization. Corresponding objective functions are denoted as $\mathcal{L}^{motion}_{2D}$ and $\mathcal{L}^{motion}_{3D}$.

\subsection{Late fusion of modalities} \label{sec:late_fus}
The 2D student and 3D student models, trained through cross-modal distillation, can be independently applied to a single sensor---either an RGB camera or LiDAR---depending on the specific application. To further enhance performance, we propose merging the predictions from both models for multi-sensor applications. The underlying assumption in our fusion method is that the pseudo-label exchange during cross-modal training should lead to consistent object regions between the two modalities. In contrast, inconsistent predictions are likely due to domain-specific noise. We therefore suggest using inter-domain consistency as a measure of confidence in the predictions. During inference, we propose a simple late fusion strategy by retaining the union of predictions from both models that overlap by at least a threshold value $\tau$, while discarding predictions \textit{unique} to only one modality.

%% file: sec/4_experiments.tex
\section{Experiments}

\subsection{Datasets}
\label{data}

We evaluate the proposed approach on TRI-PD \cite {DOM}, KITTI \cite {KITTI} and WOD \cite{waymo} datasets. \textbf{TRI-PD} is a benchmark for 2DOD, comprising an extensive collection of highly realistic synthetic videos of driving environments. The benchmark's test set contains solely RGB images. To accommodate evaluations involving a 3D model, we introduce a new test set composed of 17 scenes with 3 camera views each, randomly extracted from the former TRI-PD training set. Point clouds for each image are computed using the GT dense depth and camera poses. In all our experiments, this test set is excluded from the training sequences. The list of KITTI frames used in 3D evaluation, as well as the list of scenes of the new TRI-PD test set, are provided in the appendix. \textbf{KITTI} is a set of benchmarks designed for computer vision tasks in road scene applications. The instance segmentation subset has been adopted in previous works as a benchmark for 2DOD. This subset includes 200 frames, of which only 142 have associated 3D information (LiDAR points). We use this new subset for evaluation in the multi-modal setting. During training, all raw-data are used without labels.
\textbf{Waymo Open Dataset} (WOD) \cite{waymo} is a large-scale dataset for autonomous driving, which includes 3D point clouds and 2D RGB images. Although WOD has not been traditionally used for 2DOD benchmarks, its complex, real-world scenarios are valuable for testing our unsupervised method. For our experiments, we use point clouds from the top-mounted 64-channel LiDAR, along with video frames from the front-facing camera. The training set includes approximately 800 sequences of 200 frames each, while the validation set contains 200 sequences of 200 frames each.

\subsection{Metrics}
Consistent with previous work on object discovery \cite{DIOD}, we validate our approach using three metrics: foreground Adjusted Rand Index (fg-ARI), all-ARI, and F1@50. The \textbf{fg-ARI} measures the similarity between two clusterings by considering all pairs of points within the foreground area, counting pairs that are either assigned to the same cluster or different clusters in both the predicted and true clustering. Both metrics aim to evaluate the quality of the instance segmentation, considering only the foreground regions without relying on class labels.
The \textbf{all-ARI} is a variation of ARI that accounts for the accurate segmentation of the image background. Both of these metrics are pixel-wise measures and do not normalize for the size of the objects, which tend to be biased toward correctly segmenting larger objects. \cite{DIOD} has addressed this bias by calculating an instance-wise metric, known in object detection as \textbf{F1@50}.


\subsection{Implementation details}
\label{sec:details}
\noindent\textbf{Synthetic photo-realistic dataset (TRI-PD).} Given the availability of dense depth maps, we used the camera poses to generate XYZd-formatted input and omitted the scene completion task. Both the RGB images and front-view projections were resized to $(480 \times 968)$. Images were augmented similarly to \cite{DIOD}, while depth maps were transformed using data jittering, data drop, horizontal-flip and crop-resize, all with a probability $0.4$. The model was trained for $300$ epochs.\\
\noindent \textbf{Real-world setting (KITTI and WOD).} We forwarded RGB images to the 2D branch and front-projected 3D point clouds to the 3D branch, both using a ResNet18 \cite{resnet} backbone without pre-training. The training was conducted for $100$ epochs following a burn-in period, using batches of $8$ input sequences of length $T=5$. For each modality, the teacher parameters were computed as the EMA of the student with a keeping-rate $0.996$. For KITTI, the motion segments used for guiding the slots' learning were extracted from RAFT optical flow \cite{RAFT}, using the approach in \cite{TSAM}. For WOD, pseudo-motion segments are generated using xMOD trained on KITTI. Specific details for each branch are provided in the appendix.

\subsection{Multi-modal Object Discovery}
\label{sec:MMOD}
\input{sec/subsections/tab_allARI_f1}

In \autoref{F1}, we present the quantitative results on the three datasets for the 2D and 3D object discovery tasks. On the TRI-PD dataset the point cloud data is very dense and contains less texture compared to RGB input, simplifying the task of object discovery. Consequently, the 3DOD baseline approach (DIOD-3D) achieves significantly higher performance than the 2DOD baseline (DIOD), with a 9-point increase in F1 score. Cross-modal training further enhances the 2D model's performance by 4.9 point, attributed to the 3D model, which experiences a 2.1-point decrease mainly due to lower precision. Detailed precision and recall results are provided in the appendix. Ultimately, late fusion of modalities yields the highest performance on this dataset, achieving an F1 score of $42.5$. The sparsity of point cloud data in the KITTI and WOD datasets presents added challenges for the DIOD-3D baseline relative to the 2D baseline. Cross-modal training helps mitigate these challenges, boosting the $F1@50$ score of the 2D model by 3.6 and 7.6 points and the 3D model by $3.4$ and $4.6$ points on KITTI and WOD, respectively. In this context, late fusion proves highly beneficial, increasing performance by 5.1 points on KITTI and $7.5$ points on WOD compared to the next best model, our xMOD (2D) branch.
The discrepancy between all-ARI and F1 scores across datasets arises from the differing nature of these metrics: all-ARI is pixel-wise, while F1 score is instance-wise. This means that if the model detects a large, noisy segment, it minimally impacts the F1 score (counting as a single false positive) but lowers the all-ARI score due to many misclassified pixels. As a result, the model may perform better on TRIP-PD and Waymo in terms of F1 score, but achieve higher all-ARI on KITTI, where the effects of pixel-wise noise differ. 
\input{sec/subsections/visu_all}

\subsection{2D Object Discovery}
\input{sec/subsections/tab_fgARI}

In previous experiments, we introduced a baseline in 3DOD, which was enhanced through cross-modal training and late fusion during inference. We emphasize that these results were achieved on the new KITTI and TRI-PD test sets, with available 3D data (see section \ref{data}). In this section, for an objective comparison with previous methods in 2DOD, we evaluate the 2D branch of our model (xMOD (2D)) on the conventional test sets of the studied benchmarks. We use the most widely employed metric in 2DOD, ie. fg-ARI, for evaluation. The results in \autoref{fg-ARI} show that xMOD (2D) branch also benefits from cross-modal training, exploiting \textit{readily} available 3D data.

\subsection{Ablation studies}\label{abl}
\paragraph{Early fusion vs. late fusion.} We explored two fusion strategies for integrating RGB images and front-projected point clouds. Early fusion combines the modalities at the input level with concatenation across the channel axis, while late fusion, as explained in \autoref{sec:late_fus}, refines segmentation by cross-examining predictions from the two modalities. With an overlap threshold of $\tau=0.3$, late fusion significantly outperformed early fusion by $8$ F1 points after cross-modal training such as shown in \autoref{early_fus}.
\input{sec/subsections/tab_RGBXYZd}
\paragraph{Impact of scene completion.} We evaluated using the pretext task of scene completion, where the model estimates point positions based on neighbors. As shown in \autoref{comple}, this task helped our method discover objects, resulting in a $7.7$-point increase in F1 score.
\input{sec/subsections/tab_w-wo_completion}
\paragraph{Impact of intra-modal distillation.} Unlike prior work, we focused solely on cross-modality distillation losses, without applying intra-modality losses between the teacher and student of the same modality. Experiments showed (\autoref{losses}) that adding intra-modality losses decreased performance slightly by $0.6$ F1 points. This suggests the intra-modality loss may act as a redundant constraint, hindering the model's ability to learn valuable features from the other modality through cross-modal losses.
%
\input{sec/subsections/tab_losses}
\paragraph{Limitations on nearby and distant objects.} 
From the qualitative analysis in \autoref{fig:visu_all} and \autoref{fig:visu_3d}, we observe that segmentation quality depends on the object's distance from the camera, affecting its size in the 2D image. Based on this, we split the test set into three distance-based subsets and measure the F1 score for each in \autoref{dist}. For objects within 10 meters, which are usually cropped in images and front view projections (see the example of the red car at the top right of \autoref{fig:visu_3d}), the F1 score decreases. Mid-distance objects (10-30 meters), which are clearly visible and densely represented in the point cloud, achieve a higher F1 score. Beyond 30 meters, objects are small and LiDAR data is sparse, dropping the F1 score to 7.2 due to low recall. A potential solution is re-injecting object instances from the high-confidence range into the two other ranges to enhance model sensitivity in these areas.
\input{sec/subsections/tab_dist}
\input{sec/subsections/visu_3d}

%% file: sec/subsections/tab_allARI_f1.tex
\begin{table}
\scriptsize
\setlength\tabcolsep{1.0pt}
\centering

\begin{tblr}{
  width = \linewidth,
  colspec = {X[2.3,l,m]X[4,l,m]X[1,l,m]X[1,l,m]X[1,l,m]X[1,l,m]X[1,l,m]X[1,l,m]},
  column{1} = {rightsep=0pt},
  cells = {c},
  cell{1}{1} = {r=2}{},
  cell{1}{2} = {r=2}{},
  cell{1}{3} = {c=2}{},
  cell{1}{5} = {c=2}{},
  cell{1}{7} = {c=2}{},
  cell{3}{1} = {r=2}{},
  cell{4}{2} = {fg=blue},
  cell{5}{1} = {r=2}{},
  cell{5}{2} = {fg=blue},
  cell{6}{2} = {fg=blue},
  cell{7}{2} = {fg=blue},
  hline{1,8} = {-}{0.08em},
  hline{2} = {3-8}{},
  hline{3,5,7} = {-}{0.05em},
}
Modality & Method & TRI-PD &  & KITTI &  & WOD & \\
 &  & all-ARI & F1 & all-ARI & F1 & all-ARI & F1\\
2D & DIOD & \textbf{66.1} & 30.6 & 62.8 & 18.7 & 59.4 & 27.5\\
 & xMOD (2D) & 64.7 & 35.5 & \uline{69.7} & \uline{22.3} & \uline{66.1} & \uline{35.1}\\
3D &  DIOD-3D  & \uline{65.1} & \uline{39.6} & 51.6 & 15.5 & 55.3 & 25.6\\
 & xMOD (3D) & 65.0 & 37.5 & 58.8 & 18.9 & 62.3 & 31.0\\
Multi & xMOD (2D+3D) & 64.8 & \textbf{42.5} & \textbf{75.8} & \textbf{27.4} & \textbf{72.3} & \textbf{42.6}
\end{tblr}
\caption{\textbf{Multi-modal Object Discovery}. The models resulting from our proposed approach are presented in \textcolor{blue}{blue}. Parentheses indicate the modality used during inference. A comparison with ClusterNet \cite{wang20224d} is provided in the supplementary materials.}\label{F1}
\end{table}

%% file: sec/subsections/visu_all.tex
\begin{figure*}[tb]
\centering
\includegraphics[width=1\textwidth]{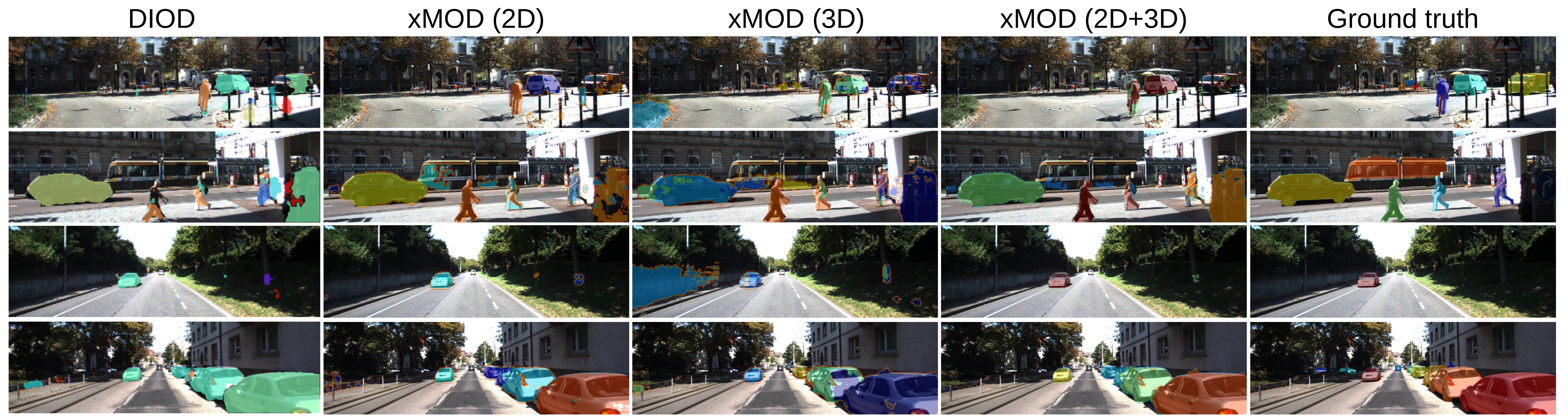}
\caption{Qualitative comparison of our method with state-of-the-art approach DIOD \cite{DIOD}, the cross-modal branches xMOD (2D), xMOD (3D) separately and the final result after fusion xMOD (2D+3D) in real-world scenes (KITTI \cite{KITTI}). Parentheses indicate the modality used during inference. Each colored mask represents the content of one slot. The segmentations are displayed above the RGB image for visualisation purposes only. Improvements in xMOD are especially evident in pedestrian detection and background noise suppression.}
\label{fig:visu_all}
\end{figure*}

%% file: sec/subsections/tab_fgARI.tex
\begin{table}[htb]
\footnotesize
\renewcommand{\arraystretch}{1.1} 
  \begin{center}
    {\scalebox{1.0}{
\begin{tabular}{clll}
\toprule  
Guidance signal & Method & TRI-PD & KITTI \\
\hline

 & DINOSAUR \cite{dinosaur} & - & 70.3 \\
\midrule
\multirow{1}{*}{optical flow}& PPMP \cite{PPMP}      & - & 51.9   \\
\midrule
flow + depth & SAVI++ \cite{SAVIpp,motok}      & - & 23.9\\
\midrule
\multirow{8}{*}{2D motion masks} & {Bao et al. \cite{DOM}} & 50.9 & 47.1 \\
& MoTok \cite{motok} & 55.1  & 64.4 \\
 & BMOD \cite{BMOD}  & 53.9 & 54.7\\
 & DIOD \cite{DIOD} & \underline{66.1} & \underline{73.5}   \\
 & xMOD (2D) & \textbf{68.0} & \textbf{75.5} \\
 \cmidrule(){2-4}
  & BMOD* \cite{BMOD}  & 58.5 & 60.8\\
  & DIOD* \cite{DIOD} & \textbf{69.7} & \underline{72.3}   \\
  & xMOD* (2D) & \underline{67.1} & \textbf{76.9} \\
\bottomrule
\end{tabular}
}}
\end{center}
\caption{Evaluation of 2D object discovery in foreground regions using fg-ARI metric on the TRI-PD and KITTI test sets. Methods using an encoder pre-trained with DINOv2 \cite{dinov2} are marked with~*. } \label{fg-ARI}
\end{table}

%% file: sec/subsections/tab_RGBXYZd.tex
\begin{table}[htb]\centering
\begin{adjustbox}{width=0.6\linewidth}
\begin{tabular}{ccc}
\toprule 
& Method & F1@50 \\
\midrule
\multirow{3}{*}{end of burn-in} & 2DOD & 9.3 \\
& 3DOD & 8.6 \\
& Early fusion & \textbf{12.8} \\
\midrule
 & Early fusion & 19.4 \\
\multirow{-2}{*}{final setting} & Late fusion & \textbf{27.4} \\
\bottomrule
\end{tabular}
\end{adjustbox}
\caption{Early \vs late fusion.} \label{early_fus}
\end{table}

%% file: sec/subsections/tab_w-wo_completion.tex
\begin{table}[htb]

\centering
\begin{adjustbox}{width=0.6\linewidth}
\begin{tabular}{ccc}
\toprule

Scene completion & all-ARI & F1@50 \\
\midrule
\xmark & 63.7 & 19.7 \\
\cmark & \textbf{75.8} & \textbf{27.4} \\

\bottomrule
\end{tabular}
\end{adjustbox}
\caption{Ablation study on the scene completion pretext task on KITTI dataset, using the late fusion strategy.} \label{comple}
\end{table}

%% file: sec/subsections/tab_losses.tex



\begin{table}[htb]
\centering
\begin{adjustbox}{width=0.55\linewidth}
\begin{tabular}{ccc}
\toprule

\multicolumn{2}{c}{Losses} & \multirow{2}{*}{F1@50}  \\
Cross-modal & Intra-modal &  \\
\midrule
\cmark & \cmark &  26.8 \\

\cmark &   &  \textbf{27.4} \\

\bottomrule
\end{tabular}
\end{adjustbox}
\caption{Analysis of the impact of intra-modal losses ($\mathcal{L}^{dist}_{2D \rightarrow 2D}$ and $\mathcal{L}^{dist}_{3D \rightarrow 3D}$) on object discovery in real-world (KITTI).} \label{losses}
\end{table}

%% file: sec/subsections/tab_dist.tex
\begin{table}[htb]
\centering
\begin{adjustbox}{width=0.8\linewidth}
\begin{tabular}{ccccc}
\toprule
Distance (m) &  AvgPts/Obj & F1@50 & Precision & Recall \\
\midrule
0-10  & 2640  & 21.7 & 68.2 & 12.9 \\
10-30 & 941  & 46.4 & 85.7 & 31.8 \\
30-70 & 134  & 7.2 & 29.5 & 4.1 \\
\midrule
0-70 & 1105  & 27.4 & 56.9 & 18.0 \\
\bottomrule
\end{tabular}
\end{adjustbox}
\caption{Object discovery performance on KITTI on 3 subsets of objects defined by their distance to the camera. AvgPts/Obj is the average number of points per object in the subset.} \label{dist}
\end{table}

%% file: sec/subsections/visu_3d.tex
\begin{figure}[tbp]
    \centering
    \begin{adjustbox}{width=1\linewidth}
    \begin{subfigure}{0.47\linewidth}
        \centering
        \includegraphics[width=1\linewidth]{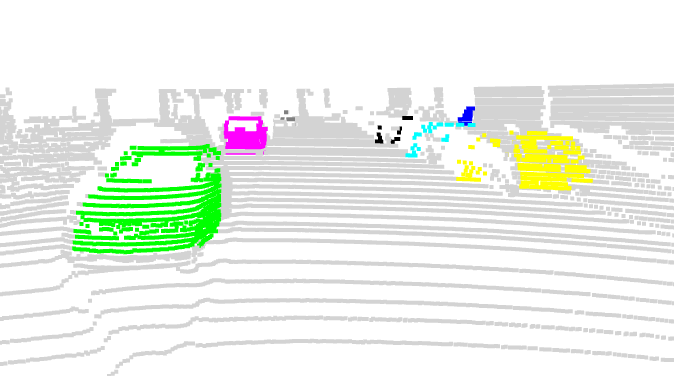}
    \end{subfigure}
    \begin{subfigure}{0.47\linewidth}
        \centering
        \includegraphics[width=1\linewidth]{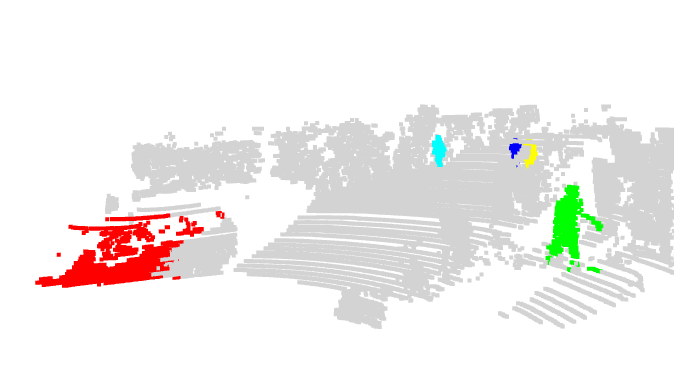}
    \end{subfigure}
    \end{adjustbox}
    \begin{adjustbox}{width=1\linewidth}
    \begin{subfigure}{0.47\linewidth}
        \centering
        \includegraphics[width=1\linewidth]{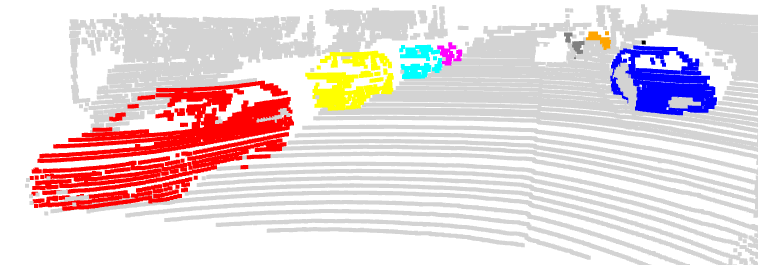}
    \end{subfigure}
    \begin{subfigure}{0.47\linewidth}
        \centering
        \includegraphics[width=1\linewidth]{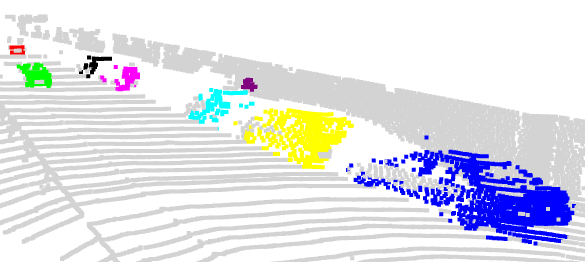}
    \end{subfigure}
    \end{adjustbox}
    
    \caption{3D visualization of predictions produced by xMOD (2D+3D). The background is displayed in gray and  each colored mask represents the content of a distinct slot.}
    \label{fig:visu_3d}
\end{figure}

%% file: sec/5_conclusion.tex
\section{Conclusion}
In this work, we first presented a method for discovering multiple objects in 3D data. Our approach builds on the latest advancements in motion-guided object discovery in images and introduces necessary adjustments to handle sparse 3D point cloud data from LiDAR sensors. In particular, we found that scene completion is a well-suited pretext task for 3DOD, as scene understanding is critical in this unsupervised setting. We also proposed a cross-modal distillation training method, where two branches, each processing a distinct modality---2D or 3D---exchange pseudo-labels during training. The experiments showed advantages for both modalities, which can be attributed to the limitations of each sensor when used independently. 
To further investigate the multi-modal setting, we proposed a late fusion strategy during inference, using multi-modal consistency as a confidence criterion. The high precision of this approach at medium distances opens perspective for instance injection methods to improve the model reliability in more challenging conditions. Future work could also explore the use of multi-scale supervision---beyond the latent space---to address the reduced sensitivity observed for small objects.

\noindent This work was supported by a French government grant managed by the Agence Nationale de la Recherche under the France 2030 program with the reference "ANR-23-DEGR-0001". This publication was made possible by the use of the FactoryIA supercomputer, financially supported by the Ile-De-France Regional Council.

%% file: supp_mat.tex
\clearpage
\setcounter{page}{1}
\maketitlesupplementary
\section{Further implementation details}
In this section, we provide further branch-specific implementation details, about the real-world setting.
\begin{itemize}[noitemsep, left=0pt]
    \item 2D Branch: Same configuration as in \cite{DIOD}, with visual augmentations only. 
    
    \item 3D Branch: The input frontal projection is resized to $(368 \times 1248)$ and min-max normalized using global dataset-wide minimum and maximum values. The 3D student model receives sequences of images augmented using \textit{data-jittering} with a probability $0.4$, while the teacher model receives the original projected image. For the scene completion task, $20\%$ of the projected points are dropped from the input and the 3D student is trained to recover them.
\end{itemize}

\section{Detailed Multi-modal Object Discovery results}
The results in \autoref{F1_supp} indicate that the 3D baseline model (DIOD-3D) exhibits relatively low recall, particularly in real-world data (KITTI), where it achieves $13.2$, highlighting challenges in object localization within sparse 3D data. Recall improves when the 3D branch is trained alongside RGB images via cross-modal training, as seen with xMOD (3D), which increases recall to $16.9$ on KITTI. This improvement suggests that the model benefits from clearer object patterns in RGB images. On the other hand, the 2D only method presents low precision due to the noise and less homogeneous textures compared to 3D data, resulting in lower precision in 2D (\eg $17.8$ for DIOD on KITTI). 2D precision is enhanced through cross-modal training (from $17.8$ to $22.8$), possibly due to inconsistencies between RGB noise and point cloud data. The proposed late fusion in xMOD (2D + 3D) significantly boosts precision while maintaining high recall, achieving a better balance between both, as reflected by higher F1 scores of $27.4$ for KITTI and $42.5$ for TRI-PD. In \autoref{F1_distances} we see that, across specific depth ranges (10-30m), the late fusion preserves high recall while maintaining strong precision, demonstrating reduced trade-offs and improved performance.

\input{sec/subsections/detail}
\section{Definition of new benchmarks with available 3D data.} \label{sec:def}
\noindent As stated in the main paper, experiments that include 3D data are evaluated on new test sets for TRI-PD and KITTI, with available 3D data. 

\begin{itemize}
    \item The test set for TRI-PD was derived from the old training set (with these scenes excluded during training). It consists of the first $17$ non-banned scenes, each including views from cameras $1$, $5$, and $6$ :
\begin{align*}
    & \texttt{scene\_000003}, \texttt{scene\_000007}, \texttt{scene\_000010}, \\
    &\texttt{scene\_000011}, \texttt{scene\_000013}, \texttt{scene\_000014}, \\
    &\texttt{scene\_000017}, \texttt{scene\_000020}, \texttt{scene\_000022}, \\
    &\texttt{scene\_000023}, \texttt{scene\_000024}, \texttt{scene\_000025}, \\
    &\texttt{scene\_000026}, \texttt{scene\_000027}, \texttt{scene\_000030}, \\
    &\texttt{scene\_000031}, \texttt{scene\_000033}
\end{align*}
    
    \item The new test-set for KITTI includes 142 scenes, selected from the original 200 scenes in the previous test set, where LiDAR data is available. The scenes names are listed below:
\begin{itemize}[label={}, leftmargin=*, itemsep=0pt, parsep=0pt]
    \item {\texttt{000163}, \texttt{000107}, \texttt{000031}, \texttt{000090}, \texttt{000053},
    \item \texttt{000069}, \texttt{000119}, \texttt{000070}, \texttt{000075}, \texttt{000043},
    \item \texttt{000042}, \texttt{000015}, \texttt{000086}, \texttt{000048}, \texttt{000011},
    \item \texttt{000109}, \texttt{000157}, \texttt{000142}, \texttt{000037}, \texttt{000033},
    \item \texttt{000019}, \texttt{000098}, \texttt{000039}, \texttt{000016}, \texttt{000145},
    \item \texttt{000114}, \texttt{000025}, \texttt{000095}, \texttt{000113}, \texttt{000010},
    \item \texttt{000076}, \texttt{000110}, \texttt{000038}, \texttt{000018}, \texttt{000160},
    \item \texttt{000044}, \texttt{000040}, \texttt{000027}, \texttt{000034}, \texttt{000045},
    \item \texttt{000029}, \texttt{000093}, \texttt{000147}, \texttt{000122}, \texttt{000128},
    \item \texttt{000067}, \texttt{000143}, \texttt{000141}, \texttt{000047}, \texttt{000002},
    \item \texttt{000052}, \texttt{000158}, \texttt{000149}, \texttt{000020}, \texttt{000079},
    \item \texttt{000032}, \texttt{000055}, \texttt{000036}, \texttt{000097}, \texttt{000074},
    \item \texttt{000051}, \texttt{000066}, \texttt{000089}, \texttt{000092}, \texttt{000009},
    \item \texttt{000077}, \texttt{000028}, \texttt{000162}, \texttt{000115}, \texttt{000124},
    \item \texttt{000085}, \texttt{000108}, \texttt{000054}, \texttt{000080}, \texttt{000123},
    \item \texttt{000126}, \texttt{000088}, \texttt{000148}, \texttt{000094}, \texttt{000159},
    \item \texttt{000132}, \texttt{000017}, \texttt{000155}, \texttt{000078}, \texttt{000072},
    \item \texttt{000105}, \texttt{000081}, \texttt{000168}, \texttt{000073}, \texttt{000116},
    \item \texttt{000164}, \texttt{000112}, \texttt{000199}, \texttt{000056}, \texttt{000106},
    \item \texttt{000050}, \texttt{000129}, \texttt{000024}, \texttt{000068}, \texttt{000169},
    \item \texttt{000059}, \texttt{000003}, \texttt{000130}, \texttt{000065}, \texttt{000146},
    \item \texttt{000064}, \texttt{000023}, \texttt{000131}, \texttt{000144}, \texttt{000117},
    \item \texttt{000013}, \texttt{000058}, \texttt{000062}, \texttt{000049}, \texttt{000012},
    \item \texttt{000121}, \texttt{000026}, \texttt{000091}, \texttt{000150}, \texttt{000041},
    \item \texttt{000071}, \texttt{000022}, \texttt{000060}, \texttt{000046}, \texttt{000096},
    \item \texttt{000030}, \texttt{000007}, \texttt{000161}, \texttt{000111}, \texttt{000118},
    \item \texttt{000084}, \texttt{000014}, \texttt{000127}, \texttt{000008}, \texttt{000063},
    \item \texttt{000125}, \texttt{000120}, \texttt{000021}, \texttt{000057}, \texttt{000035},
    \item \texttt{000061}}
\end{itemize}

\end{itemize}

\section{Comparison with Clusternet}
In the ClusterNet \cite{wang20224d} average precision (AP) computation, a subtle yet significant implementation nuance in the handling of predictions below the Intersection over Union (IoU) threshold between predicted and ground truth masks is present. The original code lacks an explicit "else" clause when evaluating instance matches, effectively omitting predictions with IoU values below the specified threshold. This results in an incomplete categorization of predictions, where instances not meeting the IoU criterion are neither classified as true positives (TP) nor false positives (FP). Consequently, the lists tracking false positives (FP), true positives (TP), and prediction scores (scores) become inconsistent, potentially introducing computational errors in metric calculations. The original paper (main or supplementary material) did not provide any precision regarding this customized AP computation. 

Furthermore, regarding the implementation of the method itself, key components such as the 3D Instance Segmentation module are absent from the repository, thus we were not able to reproduce it in order to estimate their performances with a standard AP formula. To facilitate comparison, we report in \autoref{clust} the performances of our approach with an implementation of the AP that is similar to that of ClusterNet \cite{wang20224d}. 

\input{sec/subsections/tab_clust}

%% file: sec/subsections/detail.tex
\begin{table*}[htb]
\renewcommand{\arraystretch}{1.1} 
\centering
\begin{tabular}{ccccccccc}
\toprule
\multicolumn{1}{c}{\multirow{2}{*}{}} & \multicolumn{4}{c}{KITTI} & \multicolumn{4}{c}{TRI-PD} \\ \cline{2-9} 
\multicolumn{1}{c}{} & all-ARI & F1@50 & Precision & Recall & all-ARI & F1@50 & Precision & Recall \\ \hline
DIOD \cite{DIOD} & 62.8 & 18.7 & 17.8 & \underline{19.7} & \textbf{66.1} & 30,6 & 22,4 & \textbf{48,1} \\ 
\cline{2-9} 
 \textcolor{blue}{xMOD (2D)} & \underline{69.7} & \underline{22.3} & \underline{22.8} & \textbf{21.8} & 64.7 & 35.5 & 30.4 & 42.8 \\ 
 \hline
\textcolor{blue}{DIOD-3D} & 51.6 & 15.5 & 18.9 & 13.2 & \underline{65.1} & \underline{39.6} & \textbf{47.0} & 34.3 \\ \cline{2-9} 
 \textcolor{blue}{xMOD (3D)} & 58.8 & 18.9 & 21.6 & 16.9 & 65.0 & 37.5 & 32.4 & \underline{44.6} \\
 \hline
\textcolor{blue}{xMOD (2D + 3D)} & \textbf{75.8} & \textbf{27.4} & \textbf{56.9} & 18.0 & 64.8 & \textbf{42.5} & \underline{42.9} & 42.0 \\ 
\bottomrule
\end{tabular}
\caption{\textbf{Multi-modal Object Discovery} evaluated on the new KITTI and TRI-PD testsets with available 3D data (see \autoref{sec:def}). The models resulting from our proposed approach are presented in \textcolor{blue}{blue}. Parentheses indicate the modality used during inference.} \label{F1_supp}
\end{table*}

\begin{table*}[htb]
\renewcommand{\arraystretch}{1.1} 
\centering
\begin{tabular}{cccccccccc}
\toprule
\multicolumn{1}{c}{\multirow{2}{*}{}} & \multicolumn{3}{c}{0-10} & \multicolumn{3}{c}{10-30} & \multicolumn{3}{c}{30-70} \\ \cline{2-10} 
\multicolumn{1}{c}{}  & F1@50 & Precision & Recall & F1@50 & Precision & Recall & F1@50 & Precision & Recall\\ 
\hline
DIOD \cite{DIOD} &  15.3 & 11.6 & 25.4 & 26.2 & 21.6 & 34.8 & 12.7 & 8.5 & 21.1\\ 
\cline{2-10} 
 \textcolor{blue}{xMOD (2D)} & 20.6 & 30.8 & 15.6 & 32.5 & 31.1 & 34.3 & 16.0 & 15.0 & 16.5 \\ 
 \hline
\textcolor{blue}{DIOD-3D} & 15.2 & 30.3 & 10.7 & 22.9 & 21.9 & 24.0 & 8.9 & 20.0 & 5.5 \\ 
\cline{2-10} 
 \textcolor{blue}{xMOD (3D)} & 20.2 & 60.7 & 12.3 & 28.7 & 25.2 & 32.8 & 10.1 & 9.4 & 12.1
 \\
 \hline
\textcolor{blue}{xMOD (2D + 3D)} & 21.7 & 68.2 & 12.9 & 46.4 & 85.7 & 31.8 & 7.2 & 29.5 & 4.1 \\ 
\bottomrule
\end{tabular}
\caption{\textbf{Multi-modal Object Discovery on KITTI for different subsets of object defined by their distance to the camera}. The models resulting from our proposed approach are presented in \textcolor{blue}{blue}. Parentheses indicate the modality used during inference.} \label{F1_distances}
\end{table*}

%% file: sec/subsections/tab_clust.tex
\begin{table}[tb]
\centering
\begin{tabular}{ccccc}
\toprule
\multicolumn{1}{c}{\multirow{1}{*}{Modality}} &\multicolumn{1}{c}{\multirow{1}{*}{Method}} & \multicolumn{1}{c}{\multirow{1}{*}{WOD}} \\

\midrule

 & Clusternet \cite{wang20224d} & \textbf{26.2} \\ 
\multirow{-2}{*}{Multi} & \textcolor{blue}{xMOD (2D+3D)} & \textbf{30.0} \\ 

\bottomrule
\end{tabular}
\caption{\textbf{Multi-modal Object Discovery evaluated on the dataset WOD in AP@70}. The models resulting from our proposed approach are presented in \textcolor{blue}{blue}. .} \label{clust}
\end{table}

%% file: main.bbl
\begin{thebibliography}{45}
\providecommand{\natexlab}[1]{#1}
\providecommand{\url}[1]{\texttt{#1}}
\expandafter\ifx\csname urlstyle\endcsname\relax
  \providecommand{\doi}[1]{doi: #1}\else
  \providecommand{\doi}{doi: \begingroup \urlstyle{rm}\Url}\fi

\bibitem[Bao et~al.(2022)Bao, Tokmakov, Jabri, Wang, Gaidon, and Hebert]{DOM}
Zhipeng Bao, Pavel Tokmakov, Allan Jabri, Yu-Xiong Wang, Adrien Gaidon, and
  Martial Hebert.
\newblock Discorying object that can move.
\newblock In \emph{CVPR}, 2022.

\bibitem[Bao et~al.(2023)Bao, Tokmakov, Wang, Gaidon, and Hebert]{motok}
Zhipeng Bao, Pavel Tokmakov, Yu-Xiong Wang, Adrien Gaidon, and Martial Hebert.
\newblock Object discovery from motion-guided tokens.
\newblock In \emph{CVPR}, 2023.

\bibitem[Bogoslavskyi and Stachniss(2017)]{groundRemoval}
Igor Bogoslavskyi and Cyrill Stachniss.
\newblock Efficient online segmentation for sparse 3d laser scans.
\newblock \emph{PFG--Journal of Photogrammetry, Remote Sensing and
  Geoinformation Science}, 85:\penalty0 41--52, 2017.

\bibitem[Carion et~al.(2020)Carion, Massa, Synnaeve, Usunier, Kirillov, and
  Zagoruyko]{detr}
Nicolas Carion, Francisco Massa, Gabriel Synnaeve, Nicolas Usunier, Alexander
  Kirillov, and Sergey Zagoruyko.
\newblock End-to-end object detection with transformers.
\newblock In \emph{European conference on computer vision}, pages 213--229.
  Springer, 2020.

\bibitem[Caron et~al.(2021)Caron, Touvron, Misra, J{\'e}gou, Mairal,
  Bojanowski, and Joulin]{DINO}
Mathilde Caron, Hugo Touvron, Ishan Misra, Herv{\'e} J{\'e}gou, Julien Mairal,
  Piotr Bojanowski, and Armand Joulin.
\newblock Emerging properties in self-supervised vision transformers.
\newblock In \emph{Proceedings of the IEEE/CVF International Conference on
  Computer Vision}, pages 9650--9660, 2021.

\bibitem[Cen et~al.(2021)Cen, Yun, Cai, Wang, and Liu]{paper3}
Jun Cen, Peng Yun, Junhao Cai, Michael Wang, and Ming Liu.
\newblock Open-set 3d object detection.
\newblock pages 869--878, 2021.

\bibitem[Cheng et~al.(2024)Cheng, Song, Ge, Liu, Wang, and Shan]{yoloworld}
Tianheng Cheng, Lin Song, Yixiao Ge, Wenyu Liu, Xinggang Wang, and Ying Shan.
\newblock Yolo-world: Real-time open-vocabulary object detection.
\newblock \emph{2024 IEEE/CVF Conference on Computer Vision and Pattern
  Recognition (CVPR)}, pages 16901--16911, 2024.

\bibitem[Dave et~al.(2019)Dave, Tokmakov, and Ramanan]{TSAM}
Achal Dave, Pavel Tokmakov, and Deva Ramanan.
\newblock Towards segmenting anything that moves.
\newblock In \emph{Proceedings of the IEEE/CVF International Conference on
  Computer Vision (ICCV) Workshops}, 2019.

\bibitem[Dewan et~al.(2016)Dewan, Caselitz, Tipaldi, and Burgard]{paper2}
Ayush Dewan, Tim Caselitz, Gian~Diego Tipaldi, and Wolfram Burgard.
\newblock Motion-based detection and tracking in 3d lidar scans.
\newblock In \emph{2016 IEEE international conference on robotics and
  automation (ICRA)}, pages 4508--4513. IEEE, 2016.

\bibitem[Elsayed et~al.(2022)Elsayed, Mahendran, van Steenkiste, Greff, Mozer,
  and Kipf]{SAVIpp}
Gamaleldin~F. Elsayed, Aravindh Mahendran, Sjoerd van Steenkiste, Klaus Greff,
  Michael~C. Mozer, and Thomas Kipf.
\newblock {SAVi++}: Towards end-to-end object-centric learning from real-world
  videos.
\newblock In \emph{Advances in Neural Information Processing Systems}, 2022.

\bibitem[Fruhwirth-Reisinger et~al.(2024)Fruhwirth-Reisinger, Lin, Mali'c,
  Bischof, and Possegger]{paper11}
Christian Fruhwirth-Reisinger, Wei Lin, Duvsan Mali'c, Horst Bischof, and Horst
  Possegger.
\newblock Vision-language guidance for lidar-based unsupervised 3d object
  detection.
\newblock \emph{ArXiv}, abs/2408.03790, 2024.

\bibitem[Geiger et~al.(2013)Geiger, Lenz, Stiller, and Urtasun]{KITTI}
Andreas Geiger, Philip Lenz, Christoph Stiller, and Raquel Urtasun.
\newblock Vision meets robotics: The kitti dataset.
\newblock \emph{International Journal of Robotics Research (IJRR)}, 2013.

\bibitem[He et~al.(2015)He, Zhang, Ren, and Sun]{resnet}
Kaiming He, X. Zhang, Shaoqing Ren, and Jian Sun.
\newblock Deep residual learning for image recognition.
\newblock \emph{2016 IEEE Conference on Computer Vision and Pattern Recognition
  (CVPR)}, pages 770--778, 2015.

\bibitem[Kara et~al.(2024{\natexlab{a}})Kara, Ammar, Chabot, and Pham]{BMOD}
Sandra Kara, Hejer Ammar, Florian Chabot, and Quoc-Cuong Pham.
\newblock The background also matters: Background-aware motion-guided objects
  discovery.
\newblock In \emph{Proceedings of the IEEE/CVF Winter Conference on
  Applications of Computer Vision}, pages 1216--1225, 2024{\natexlab{a}}.

\bibitem[Kara et~al.(2024{\natexlab{b}})Kara, Ammar, Denize, Chabot, and
  Pham]{DIOD}
Sandra Kara, Hejer Ammar, Julien Denize, Florian Chabot, and Quoc-Cuong Pham.
\newblock Diod: Self-distillation meets object discovery.
\newblock In \emph{Proceedings of the IEEE/CVF Conference on Computer Vision
  and Pattern Recognition}, pages 3975--3985, 2024{\natexlab{b}}.

\bibitem[Karazija et~al.(2022)Karazija, Choudhury, Laina, Rupprecht, and
  Vedaldi]{PPMP}
Laurynas Karazija, Subhabrata Choudhury, Iro Laina, Christian Rupprecht, and
  Andrea Vedaldi.
\newblock Unsupervised multi-object segmentation by predicting probable motion
  patterns.
\newblock \emph{Advances in Neural Information Processing Systems},
  35:\penalty0 2128--2141, 2022.

\bibitem[Kipf et~al.(2022)Kipf, Elsayed, Mahendran, Stone, Sabour, Heigold,
  Jonschkowski, Dosovitskiy, and Greff]{SAVI}
Thomas Kipf, Gamaleldin~F. Elsayed, Aravindh Mahendran, Austin Stone, Sara
  Sabour, Georg Heigold, Rico Jonschkowski, Alexey Dosovitskiy, and Klaus
  Greff.
\newblock {Conditional Object-Centric Learning from Video}.
\newblock In \emph{International Conference on Learning Representations
  (ICLR)}, 2022.

\bibitem[Lin and Caesar(2024)]{ICP}
Yancong Lin and Holger Caesar.
\newblock Icp-flow: Lidar scene flow estimation with icp.
\newblock In \emph{Proceedings of the IEEE/CVF Conference on Computer Vision
  and Pattern Recognition}, pages 15501--15511, 2024.

\bibitem[Liu et~al.(2023)Liu, Zeng, Ren, Li, Zhang, Yang, Li, Yang, Su, Zhu,
  et~al.]{groundingdino}
Shilong Liu, Zhaoyang Zeng, Tianhe Ren, Feng Li, Hao Zhang, Jie Yang, Chunyuan
  Li, Jianwei Yang, Hang Su, Jun Zhu, et~al.
\newblock Grounding dino: Marrying dino with grounded pre-training for open-set
  object detection.
\newblock \emph{arXiv preprint arXiv:2303.05499}, 2023.

\bibitem[Liu et~al.(2019)Liu, Qi, and Guibas]{FN3D}
Xingyu Liu, Charles~R Qi, and Leonidas~J Guibas.
\newblock Flownet3d: Learning scene flow in 3d point clouds.
\newblock \emph{CVPR}, 2019.

\bibitem[Locatello et~al.(2020)Locatello, Weissenborn, Unterthiner, Mahendran,
  Heigold, Uszkoreit, Dosovitskiy, and Kipf]{SA}
Francesco Locatello, Dirk Weissenborn, Thomas Unterthiner, Aravindh Mahendran,
  Georg Heigold, Jakob Uszkoreit, Alexey Dosovitskiy, and Thomas Kipf.
\newblock Object-centric learning with slot attention.
\newblock In \emph{Advances in Neural Information Processing Systems}, pages
  11525--11538. Curran Associates, Inc., 2020.

\bibitem[Luo et~al.(2023)Luo, Liu, Chen, You, Benaim, Phoo, Campbell, Sun,
  Hariharan, and Weinberger]{paper7}
Katie Luo, Zhenzhen Liu, Xiangyu Chen, Yurong You, Sagie Benaim, Cheng~Perng
  Phoo, Mark Campbell, Wen Sun, Bharath Hariharan, and Kilian~Q Weinberger.
\newblock Reward finetuning for faster and more accurate unsupervised object
  discovery.
\newblock \emph{Advances in Neural Information Processing Systems},
  36:\penalty0 13250--13266, 2023.

\bibitem[Mittal et~al.(2020)Mittal, Okorn, and Held]{MIT}
Himangi Mittal, Brian Okorn, and David Held.
\newblock Just go with the flow: Self-supervised scene flow estimation.
\newblock In \emph{Proceedings of the IEEE/CVF Conference on Computer Vision
  and Pattern Recognition (CVPR)}, 2020.

\bibitem[Najibi et~al.(2022)Najibi, Ji, Zhou, Qi, Yan, Ettinger, and
  Anguelov]{paper5}
Mahyar Najibi, Jingwei Ji, Yin Zhou, Charles~R Qi, Xinchen Yan, Scott Ettinger,
  and Dragomir Anguelov.
\newblock Motion inspired unsupervised perception and prediction in autonomous
  driving.
\newblock In \emph{European Conference on Computer Vision}, pages 424--443.
  Springer, 2022.

\bibitem[Oquab et~al.(2023)Oquab, Darcet, Moutakanni, Vo, Szafraniec, Khalidov,
  Fernandez, Haziza, Massa, El-Nouby, Assran, Ballas, Galuba, Howes, Huang, Li,
  Misra, Rabbat, Sharma, Synnaeve, Xu, Jegou, Mairal, Labatut, Joulin, and
  Bojanowski]{dinov2}
Maxime Oquab, Timothée Darcet, Théo Moutakanni, Huy Vo, Marc Szafraniec,
  Vasil Khalidov, Pierre Fernandez, Daniel Haziza, Francisco Massa, Alaaeldin
  El-Nouby, Mahmoud Assran, Nicolas Ballas, Wojciech Galuba, Russell Howes,
  Po-Yao Huang, Shang-Wen Li, Ishan Misra, Michael Rabbat, Vasu Sharma, Gabriel
  Synnaeve, Hu Xu, Hervé Jegou, Julien Mairal, Patrick Labatut, Armand Joulin,
  and Piotr Bojanowski.
\newblock Dinov2: Learning robust visual features without supervision, 2023.

\bibitem[Redmon et~al.(2016)Redmon, Divvala, Girshick, and Farhadi]{yolo}
Joseph Redmon, Santosh Divvala, Ross Girshick, and Ali Farhadi.
\newblock You only look once: Unified, real-time object detection.
\newblock In \emph{Proceedings of the IEEE Conference on Computer Vision and
  Pattern Recognition (CVPR)}, 2016.

\bibitem[Ren et~al.(2015)Ren, He, Girshick, and Sun]{faster}
Shaoqing Ren, Kaiming He, Ross Girshick, and Jian Sun.
\newblock Faster r-cnn: Towards real-time object detection with region proposal
  networks.
\newblock In \emph{Advances in Neural Information Processing Systems}. Curran
  Associates, Inc., 2015.

\bibitem[Safadoust and G\"uney(2023)]{safadoust}
Sadra Safadoust and Fatma G\"uney.
\newblock Multi-object discovery by low-dimensional object motion.
\newblock In \emph{ICCV}, pages 734--744, 2023.

\bibitem[Seidenschwarz et~al.(2024)Seidenschwarz, Osep, Ferroni, Lucey, and
  Leal-Taix{\'e}]{semoli}
Jenny Seidenschwarz, Aljosa Osep, Francesco Ferroni, Simon Lucey, and Laura
  Leal-Taix{\'e}.
\newblock Semoli: What moves together belongs together.
\newblock In \emph{Proceedings of the IEEE/CVF Conference on Computer Vision
  and Pattern Recognition}, pages 14685--14694, 2024.

\bibitem[Seitzer et~al.(2022)Seitzer, Horn, Zadaianchuk, Zietlow, Xiao,
  Simon-Gabriel, He, Zhang, Scholkopf, Brox, and Locatello]{dinosaur}
Maximilian Seitzer, Max Horn, Andrii Zadaianchuk, Dominik Zietlow, Tianjun
  Xiao, Carl-Johann Simon-Gabriel, Tong He, Zheng Zhang, Bernhard Scholkopf,
  Thomas Brox, and Francesco Locatello.
\newblock Bridging the gap to real-world object-centric learning.
\newblock \emph{ArXiv}, abs/2209.14860, 2022.

\bibitem[Sim'eoni et~al.(2021)Sim'eoni, Puy, Vo, Roburin, Gidaris, Bursuc,
  P'erez, Marlet, and Ponce]{LOST}
Oriane Sim'eoni, Gilles Puy, Huy~V. Vo, Simon Roburin, Spyros Gidaris, Andrei
  Bursuc, Patrick P'erez, Renaud Marlet, and Jean Ponce.
\newblock Localizing objects with self-supervised transformers and no labels.
\newblock In \emph{BMVC}, 2021.

\bibitem[Sim\'eoni et~al.(2024)Sim\'eoni, Zablocki, Gidaris, Puy, and
  P\'erez]{survey}
Oriane Sim\'eoni, {\'{E}}loi Zablocki, Spyros Gidaris, Gilles Puy, and Patrick
  P\'erez.
\newblock Unsupervised object localization in the era of self-supervised vits:
  A survey.
\newblock In \emph{IJCV}, 2024.

\bibitem[Sun et~al.(2018)Sun, Yang, Liu, and Kautz]{PWC}
Deqing Sun, Xiaodong Yang, Ming-Yu Liu, and Jan Kautz.
\newblock {PWC-Net}: {CNNs} for optical flow using pyramid, warping, and cost
  volume.
\newblock In \emph{CVPR}, 2018.

\bibitem[Sun et~al.(2020)Sun, Kretzschmar, Dotiwalla, Chouard, Patnaik, Tsui,
  Guo, Zhou, Chai, Caine, et~al.]{waymo}
Pei Sun, Henrik Kretzschmar, Xerxes Dotiwalla, Aurelien Chouard, Vijaysai
  Patnaik, Paul Tsui, James Guo, Yin Zhou, Yuning Chai, Benjamin Caine, et~al.
\newblock Scalability in perception for autonomous driving: Waymo open dataset.
\newblock In \emph{Proceedings of the IEEE/CVF conference on computer vision
  and pattern recognition}, pages 2446--2454, 2020.

\bibitem[Teed and Deng(2020)]{RAFT}
Zachary Teed and Jia Deng.
\newblock Raft: Recurrent all-pairs field transforms for optical flow.
\newblock In \emph{Computer Vision--ECCV 2020: 16th European Conference,
  Glasgow, UK, August 23--28, 2020, Proceedings, Part II 16}, pages 402--419.
  Springer, 2020.

\bibitem[Vaswani et~al.(2017)Vaswani, Shazeer, Parmar, Uszkoreit, Jones, Gomez,
  Kaiser, and Polosukhin]{attention}
Ashish Vaswani, Noam Shazeer, Niki Parmar, Jakob Uszkoreit, Llion Jones,
  Aidan~N Gomez, \L~ukasz Kaiser, and Illia Polosukhin.
\newblock Attention is all you need.
\newblock In \emph{Advances in Neural Information Processing Systems}. Curran
  Associates, Inc., 2017.

\bibitem[Wang et~al.(2022{\natexlab{a}})Wang, Chen, and ZHANG]{wang20224d}
Yuqi Wang, Yuntao Chen, and ZHAO-XIANG ZHANG.
\newblock 4d unsupervised object discovery.
\newblock \emph{Advances in Neural Information Processing Systems},
  35:\penalty0 35563--35575, 2022{\natexlab{a}}.

\bibitem[Wang et~al.(2022{\natexlab{b}})Wang, Shen, Hu, Yuan, Crowley, and
  Vaufreydaz]{TokenCut}
Yangtao Wang, Xi Shen, Shell~Xu Hu, Yuan Yuan, James~L Crowley, and Dominique
  Vaufreydaz.
\newblock Self-supervised transformers for unsupervised object discovery using
  normalized cut.
\newblock In \emph{Proceedings of the IEEE/CVF Conference on Computer Vision
  and Pattern Recognition}, pages 14543--14553, 2022{\natexlab{b}}.

\bibitem[Wang et~al.(2023)Wang, He, Peng, Lin, Bao, and
  Zhou]{wang2023autorecon}
Yuang Wang, Xingyi He, Sida Peng, Haotong Lin, Hujun Bao, and Xiaowei Zhou.
\newblock Autorecon: Automated 3d object discovery and reconstruction.
\newblock In \emph{Proceedings of the IEEE/CVF Conference on Computer Vision
  and Pattern Recognition}, pages 21382--21391, 2023.

\bibitem[Wang et~al.(2024)Wang, Lipson, and Deng]{SEARAFT}
Yihan Wang, Lahav Lipson, and Jia Deng.
\newblock Sea-raft: Simple, efficient, accurate raft for optical flow.
\newblock \emph{arXiv preprint arXiv:2405.14793}, 2024.

\bibitem[Xie et~al.(2022)Xie, Xie, and Zisserman]{layerSeg}
Junyu Xie, Weidi Xie, and Andrew Zisserman.
\newblock Segmenting moving objects via an object-centric layered
  representation.
\newblock In \emph{Advances in Neural Information Processing Systems}, 2022.

\bibitem[Yang et~al.(2021)Yang, Lamdouar, Lu, Zisserman, and Xie]{motionGroup}
Charig Yang, Hala Lamdouar, Erika Lu, Andrew Zisserman, and Weidi Xie.
\newblock Self-supervised video object segmentation by motion grouping.
\newblock In \emph{ICCV}, 2021.

\bibitem[You et~al.(2022)You, Luo, Phoo, Chao, Sun, Hariharan, Campbell, and
  Weinberger]{paper6}
Yurong You, Katie Luo, Cheng~Perng Phoo, Wei-Lun Chao, Wen Sun, Bharath
  Hariharan, Mark~E. Campbell, and Kilian~Q. Weinberger.
\newblock Learning to detect mobile objects from lidar scans without labels.
\newblock \emph{2022 IEEE/CVF Conference on Computer Vision and Pattern
  Recognition (CVPR)}, pages 1120--1130, 2022.

\bibitem[Zadaianchuk et~al.(2023)Zadaianchuk, Seitzer, and Martius]{videosaur}
Andrii Zadaianchuk, Maximilian Seitzer, and Georg Martius.
\newblock Object-centric learning for real-world videos by predicting temporal
  feature similarities.
\newblock In \emph{NeurIPS}, 2023.

\bibitem[Zhang et~al.(2023)Zhang, Yang, Xiong, Casas, Yang, Ren, and
  Urtasun]{paper8}
Lunjun Zhang, Anqi~Joyce Yang, Yuwen Xiong, Sergio Casas, Bin Yang, Mengye Ren,
  and Raquel Urtasun.
\newblock Towards unsupervised object detection from lidar point clouds.
\newblock In \emph{Proceedings of the IEEE/CVF Conference on Computer Vision
  and Pattern Recognition}, pages 9317--9328, 2023.

\end{thebibliography}
